\documentclass[runningheads]{llncs}
\usepackage[T1]{fontenc}
\usepackage{graphicx}
\usepackage{comment}
\usepackage{amsmath}
\usepackage{booktabs}
\usepackage{algorithm}
\usepackage{algpseudocode}
\algrenewcommand\algorithmicrequire{\textbf{Input:}}
\algrenewcommand\algorithmicensure{\textbf{Output:}}
\usepackage{subcaption}
\usepackage{multirow}

\makeatletter
\newcommand{\printfnsymbol}[1]{%
\textsuperscript{\@fnsymbol{#1}}%
}
\makeatother
\begin{document}
\title{Clustering-friendly Representation Learning for Enhancing Salient Features}

\author{Toshiyuki Oshima\thanks{ First two authors have equal contribution.} \and
    Kentaro Takagi\printfnsymbol{1} \and
    Kouta Nakata}

\authorrunning{T. Oshima et al.}

\institute{Corporate R\&D Center, Toshiba
    1, Komukai Toshiba-cho, Saiwai-ku, Kawasaki, Kanagawa, Japan
    \email{\{toshiyuki1.oshima,kentaro1.takagi,kouta.nakata\}@toshiba.co.jp}}

\maketitle

\begin{abstract}
    Recently, representation learning with contrastive learning algorithms has been successfully applied to challenging unlabeled datasets.
    However, these methods are unable to distinguish important features from unimportant ones under simply unsupervised settings, and definitions of importance vary according to the type of downstream task or analysis goal, such as the identification of objects or backgrounds.
    In this paper, we focus on unsupervised image clustering as the downstream task and propose a representation learning method that enhances features critical to the clustering task.
    We extend a clustering-friendly contrastive learning method and incorporate a \textit{contrastive analysis} approach, which utilizes a reference dataset to separate important features from unimportant ones, into the design of loss functions.
    Conducting an experimental evaluation of image clustering for three datasets with characteristic backgrounds, we show that for all datasets, our method achieves higher clustering scores compared with conventional contrastive analysis and deep clustering methods.
\end{abstract}
\section{Introduction}
Clustering is one of the most fundamental methods for unsupervised machine learning and it aims to classify objects based on measures of similarity between unlabeled samples.
Advancements in deep learning techniques have produced \textit{deep clustering} methods, in which feature extraction with deep neural networks (DNNs) and a clustering process are integrated at various levels~\cite{zhou2022comprehensive}.
Self-supervised learning (SSL) has also recently been attracting attention in the representation learning field.
SSL methods learn representations by solving user-defined pretext tasks, such as instance discrimination tasks and contrastive learning, and their impressive performance in capturing visual features has been demonstrated even for complex real-world datasets~\cite{chen_simple_2020,wu_unsupervised_2018,caron_unsupervised_2020,chen_simsiam_2021}.
\textit{High-resolution} representations extracted by SSL are naturally expected to be applied to clustering.
Several approaches to simultaneously performing SSL and clustering tasks have been proposed and have achieved state-of-the-art performance~\cite{niu_spice_2022,li_contrastive_2021,tao_clustering-friendly_2022,gansbake_scan_2020}.

\begin{figure*}[t]
    \centering
    \includegraphics[width=8cm]{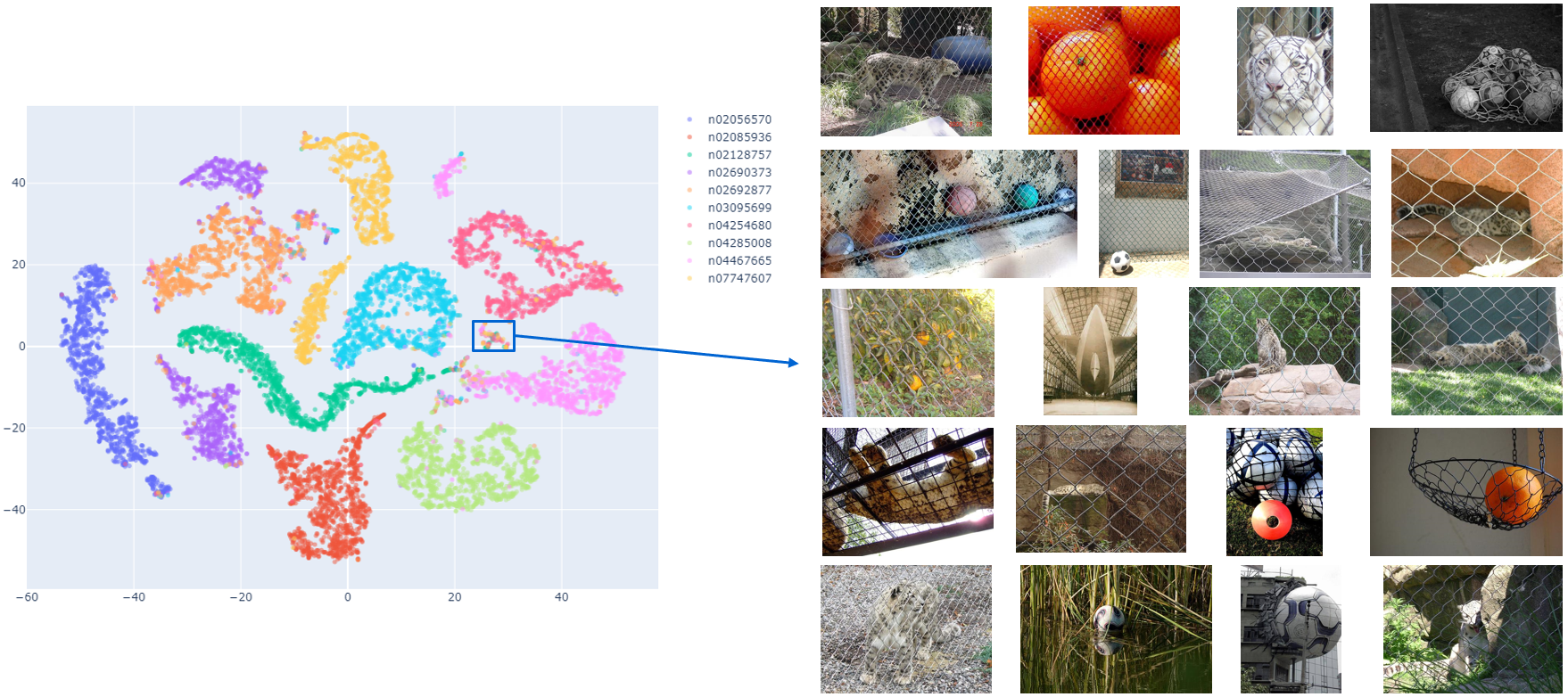}
    \caption{Distribution of feature representations on ImageNet-10 learned by IDFD (Our reproduction of Fig. 6 in \cite{tao_clustering-friendly_2022}). Some samples are grouped according to the features of the mesh structure.}\label{fig:stl10}
\end{figure*}

However, simply capturing features at high resolution or in ways that are easy for machines to understand does not necessarily improve clustering scores, and can even worsen them,
due to differences in human and algorithmic criteria for feature importance.
As a concrete example, Fig.~\ref{fig:stl10} shows instance discrimination and feature decorrelation (IDFD), an SSL method, forming a cluster depending on features of a prominent mesh structure (foreground component) rather than objects such as a leopard or orange.
Even though each feature is correctly extracted, this is not a desirable situation in the context of clustering that focuses on objects.
Similar problems can arise when the methods are applied to real-world problems. In the case of inspection images taken at a factory, for example, it is not uncommon for complex product patterns to become more prominent than the object defects targeted for classification \cite{mizukami}.
For this reason, models require proper guidance so that only features that are important for the targeted clustering are picked up.

Contrastive analysis (CA)~\cite{ge_rca_2016} is one technique for providing inductive bias with unsupervised methods to distinguish important from unimportant components.
In CA settings, two datasets are prepared: a target dataset that includes salient features of interest for clustering and a background dataset that contains only information that is ineffectual for clustering and should be discarded.
Several unsupervised methods making use of CA settings have recently been proposed~\cite{abid_exploring_2018,dirie_contrastive_2019,abid_contrastive_2019}.
These architectures have successfully isolated and extracted variables of interest from backgrounds, but they use classical architectures and learning schemes such as matrix decomposition and simple encoder-decoder structures, and they mainly minimize reconstruction loss functions.
There is thus room for improvement in the ability to learn representations for more complex real-world datasets.

In this paper, we propose contrastive IDFD (cIDFD), a method that combines a CA setting with the latest SSL scheme based on instance discrimination.
cIDFD learns unimportant features at high resolutions from the background dataset via normal instance discrimination and feature decorrelation.
Similarities among unimportant features in target samples are also obtained and utilized to guide networks toward extracting only features that are important for desirable clustering.
This feature extraction is performed by minimizing a newly defined loss function, the contrastive instance discriminative loss function.
We adapt our method to various datasets with characteristic background patterns. Our method shows improvements over conventional methods in terms of evaluating clustering tasks.

Our main contributions are as follows:
\begin{itemize}
    \item We propose a new clustering method based on instance discrimination with a CA
          setting that selectively extracts features meaningful for the
          user-defined clustering objective.
    \item We conduct experimental evaluations of clustering for three challenging
          datasets with characteristic backgrounds. The results show that cIDFD
          achieves higher clustering scores than conventional contrastive analysis models and state-of-the-art SSL methods.
\end{itemize}

\section{Related Works}
There have recently been proposals of contrastive learning as pretext tasks for SSL, in which positive pairs generated by data augmentation from the same data are brought closer together, and negative samples generated from different data are kept apart~\cite{wu_unsupervised_2018,chen_simple_2020}.
Features learned by the pretext tasks have high generalization and achieve excellent performance on a variety of downstream tasks.
Although negative pairs played an important role in avoiding collapsing solutions, there are some drawbacks, such as the need to increase batch size, and thus several methods that do not use negative pairs have been proposed \cite{chen_simsiam_2021,caron_unsupervised_2020}.

Deep clustering has shown superior performance in a variety of areas.
End-to-end learning was proposed to simultaneously perform representation learning and clustering~\cite{zhou2022comprehensive}.
To improve representation learning ability, recent works integrate clustering and SSL, in particular contrastive learning~\cite{gansbake_scan_2020,li_contrastive_2021,mice,tao_clustering-friendly_2022,park2021improving,Dang_2021_CVPR,niu_spice_2022,li2022tcl}.
IDFD~\cite{tao_clustering-friendly_2022} focuses on the representation learning phase and proposes a clustering-friendly representation learning method that uses instance discrimination loss and a proposed feature decorrelation loss motivated by the properties of classical spectral clustering.
They achieved high performance, despite using $k$-means in the clustering phase.
MiCE~\cite{mice} and CC~\cite{li_contrastive_2021} focused on end-to-end learning by integrating contrastive learning and clustering.
SCAN~\cite{gansbake_scan_2020}, RUC~\cite{park2021improving}, NNM~\cite{Dang_2021_CVPR}, TCL\cite{li2022tcl}, and SPICE~\cite{niu_spice_2022} focused on the clustering phase and are multi-stage deep clustering methods.

The principles of contrastive analysis~\cite{ge_rca_2016} were proposed as a method to separate features to be emphasized from irrelevant features that should be suppressed.
cPCA~\cite{abid_exploring_2018} utilizes that principle to separate the principal components to be emphasized from those that should be suppressed, introducing a dataset without features to be emphasized.
Contrastive singular spectrum analysis (cSSA)~\cite{dirie_contrastive_2019} extends the cPCA concept to analyze time-series datasets.
Contrastive VAE (cVAE)~\cite{abid_contrastive_2019} was also developed to understand complex, nonlinear relations between latent variables and inputs.

\section{Proposed Method}
Our goal is to learn only those representations that are appropriate to clustering for a target dataset $\mathcal{X}=\{ x_{i} \}_{i=1}^{N}$ and cluster these samples into meaningful groups under conditions where a background dataset $\mathcal{B}=\{ b_{i} \}_{i=1}^{M}$ can be prepared.
cIDFD utilizes a dataset $\mathcal{B}$ to reject the influences of background features that are unimportant with respect to clustering for the target dataset $\mathcal{X}$.
In this section, we describe the model architecture, loss computation, and actual training process of cIDFD.
\begin{figure*}[t]
    \centering
    \includegraphics[width=10cm]{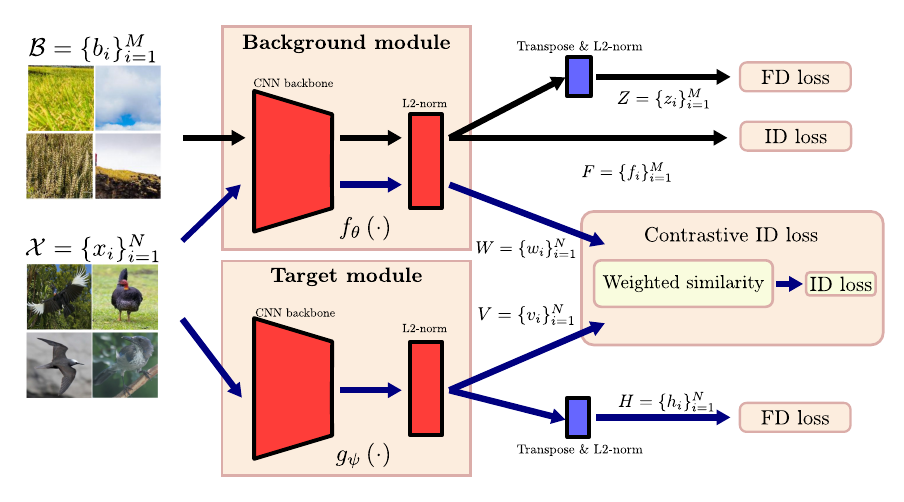}
    \caption{cIDFD framework.}\label{fig:structure}
\end{figure*}
\begin{figure*}[t]
    \begin{minipage}[b]{0.5\linewidth}
        \centering
        \includegraphics[keepaspectratio, scale=0.4]{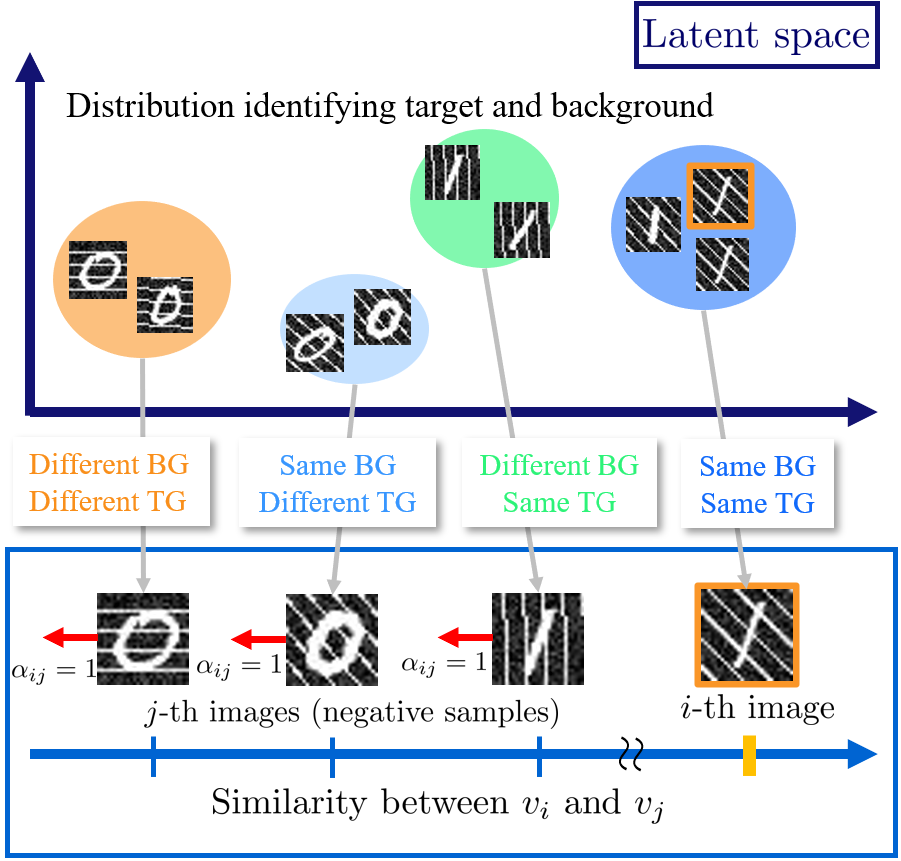}
        \subcaption{Normal instance discrimination}\label{fig:concept_id}
    \end{minipage}
    \begin{minipage}[b]{0.5\linewidth}
        \centering
        \includegraphics[keepaspectratio, scale=0.4]{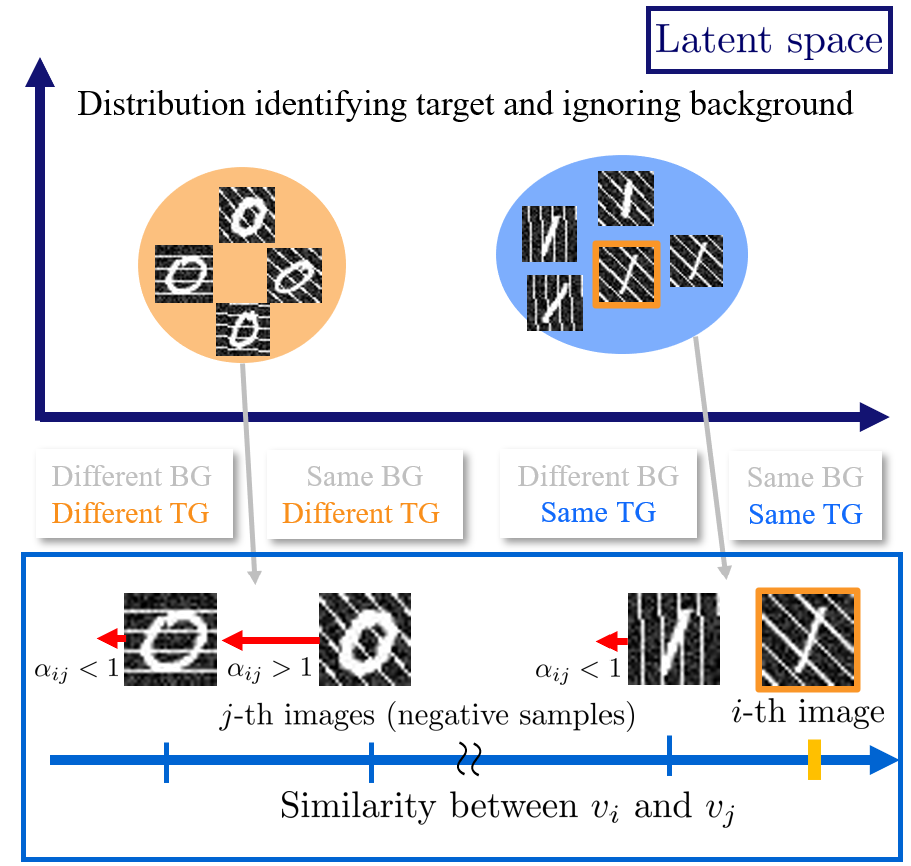}
        \subcaption{Contrastive instance discrimination}\label{fig:concept_cid}
    \end{minipage}
    \caption{Illustration of the problem in normal instance discrimination (a) and the contrastive instance discrimination strategy (b) for datasets with characteristic background patterns. Here, the number is called target (TG), the pattern of lines is called background (BG) for Stripe MNIST.
        The red arrows denote the strength of the repulsive force according to $\alpha_{ij}$ for the $i$-th and $j$-th images.}\label{fig:concept}
\end{figure*}

\subsection{The framework of cIDFD}
As Fig.~\ref{fig:structure} shows, we use two embedding functions, $f_{\theta}$ and $g_{\psi}$,
which map images to feature vectors distributed on a $d$-dimensional unit sphere.
These functions are modeled as deep neural networks with parameters $\theta$ and $\psi$, which are typically a CNN backbone and an L2-normalization layer.
$f_{\theta}$ is learned from background dataset $\mathcal{B}$ and assigned the role of extracting background features $W=\{w_{i}\}_{i=1}^{N}$ and $Z=\{z_{i}\}_{i=1}^{M}$ for image samples from $\mathcal{X}$ and $\mathcal{B}$, respectively.
The other embedding function $g_{\psi}$ is learned to extract target features $V=\{v_{i}\}_{i=1}^{N}$ from $\mathcal{X}$.
To train the background branch $f_{\theta}$, we conduct minimizing instance discrimination and feature decorrelation loss, following the same process as in a previous work~\cite{tao_clustering-friendly_2022}.
For target branch $g_{\psi}$, parameters are optimized by a newly defined loss function, contrastive instance discrimination loss, and feature decorrelation loss.
In the computation of contrastive instance discrimination loss, weighted similarity between negative samples, which includes influences from $W$ and $V$, is used as input to a non-parametric softmax classifier.

\subsection{Loss for background feature extraction}
We apply the instance discrimination proposed in~\cite{wu_unsupervised_2018} to learn
representations from a background dataset.
For given samples $\{ b_{i} \}_{i=1}^{M}$, the corresponding representations are $\{ z_{i} \}_{i=1}^{M}$ with $z_{i}=f_{\mathbf{\theta}}(b_{i})$, where $z_{i}$ is normalized to $\| z_{i}\| = 1$.
The probability of representation $z$ being assigned to the $i$th class is given by the non-parametric softmax formulation
\begin{eqnarray}
    P\left(
    i | z
    \right)
    =
    \frac{\exp(z^{T}_{i}z/\tau_{b})}
    {\sum_{j=1}^{M} \exp(z^{T}_{j}z/\tau_{b})},
\end{eqnarray}
where the dot product $z^{T}_{i}z$ is how well $z$ matches the $i$th class and $\tau_{b}$ is a temperature parameter that determines the concentration of distribution.
The learning objective is to maximize the joint probability $\prod_{i=1}^{M} P\left( i | f_{\mathbf{\theta}}(b_{i}) \right)$ as
\begin{eqnarray}
    \label{eq:loss_ib}
    \mathcal{L}_{I,b}
    = - \sum_{i=1}^{M} \log \left(
    \frac{\exp(z^{T}_{i}z_{i}/\tau_{b})}{\sum_{j=1}^{m} \exp(z^{T}_{j}z_{i}/\tau_{b})}
    \right).
\end{eqnarray}

We use a constraint for orthogonal features proposed in~\cite{tao_clustering-friendly_2022}.
The objective is to minimize
\begin{eqnarray}
    \label{eq:loss_fd}
    \mathcal{L}_{F}(F)=\mathcal{L}_{F,b}
    = - \sum_{l=1}^{d} \log\left(
    \frac{\exp(f^{T}_{l}f/\tau_{2})}
    {\sum_{j=1}^{d} \exp(f^{T}_{j}f/\tau_{2})}\right),
\end{eqnarray}
where $F=\{ f_{l} \}_{l=1}^{d}$ are latent feature vectors defined by the transpose of the latent vectors $Z$, $d$ is the dimensionality of the representations, and $\tau_{2}$ is the temperature parameter.

\subsection{Loss for target feature extraction}
\label{sec:method_ci}
When normal instance discrimination loss is used, the separation of negative pairs is performed equally for all negative pairs.
However, this makes grouping impossible, except in cases where both the target and background features are extremely similar.
Fig.~\ref{fig:concept_id} shows a conceptual illustration of such a situation.
Our motivation is to design a loss function that separates the pairs having the same background but different target features while attracting pairs having the same target features, independent of the background. For example, in Figure~\ref{fig:concept_id}, the former corresponds to the pairs of \textit{zeros} with diagonal background lines and \textit{ones} with the same diagonal background lines, while the latter corresponds to the pairs of \textit{ones}, independent of background types.

This loss function is realized by introducing weight coefficients depending on pairwise similarity between the background features of target samples $\{w_{i}\}_{i=1}^{N}$.
Our new loss function, namely, contrastive instance discrimination loss, is defined as
\begin{eqnarray}
    \label{eq:loss_ci}
    \mathcal{L}_{CI}
    = - \sum_{i=1}^{N} \log \left(
    \frac{\alpha_{ii} \exp(v_{i}^{T}v_{i}/\tau_{x})}
    {\sum_{j=1}^{N} \alpha_{ij} \exp(v_{i}^{T}v_{j}/\tau_{x}) }
    \right),
\end{eqnarray}
where $\alpha_{ij}$ is the weight coefficient between the $i$-th and $j$-th samples that determine how strongly target features $v_{i}$ and $v_{j}$ are pulled apart in the learning process.
This coefficient is formulated simply as
\begin{eqnarray}
    \label{eq:weight_coef}
    \alpha_{ij}
    = \exp(w_{i}^{T}w_{j}/\tau_{xb}).
\end{eqnarray}
When the background features of the $i$-th and $j$-th images are similar, $\alpha_{ij}$ becomes large, causing their repulsive force in the target feature space to increase.
This effect, shown in Fig.~\ref{fig:concept_cid}, reduces the similarities between the feature vectors of \textit{zeros} with diagonal lines and \textit{ones} with diagonal lines.
$\alpha_{ij}$ also becomes large for pairs having the same background and target, but there is also the usual contrastive learning effect, and their repulsion becomes weakened.
For pairs of small $\alpha_{ij}$, the repulsive force is relatively weak, and thus samples with the same target and different backgrounds are weakly separated.
We experimentally demonstrated that our method works according to the perspectives described above; the details are given in Section~\ref{sec:demo}.

Temperature parameters $\tau_{x}$ control the distribution of feature vectors, and $\tau_{xb}$ controls the magnitude of the weight coefficient $\alpha_{ij}$.
The differences from IDFD are the background module $f_\theta$ and $\alpha_{ij}$. In the case of $\lim_{\tau_{xb} \to \infty} \alpha_{ij} = 1$, cIDFD is consistent with IDFD because the contribution from the background module $f_\theta$ to the target module $g_\phi$ disappears.

\subsection{Two-stage learning}\label{sec:two-stage}
We consider separately learning the embedding functions $f_{\theta}$ and $g_{\psi}$.
In the first step, the $f_{\theta}$ branch learns a background dataset $\mathcal{B}$ by minimizing the objective function
\begin{eqnarray}
    \label{eq:loss_bg}
    \mathcal{L}_{bg} = \mathcal{L}_{I,b} + \mathcal{L}_{F}(F).
\end{eqnarray}
During this step, the input samples are only from $\mathcal{B}$, and the network parameters of $g_{\psi}$ are not updated.
After learning, $f_{\theta}$ works as an extractor of features to be discarded.
In the second step, we freeze the parameters of $f_{\theta}$ and train $g_{\psi}$ by dataset $\mathcal{X}$.
The objective function is a composition of the contrastive instance discriminative loss \eqref{eq:loss_ci} and feature decorrelation loss for the target features,
\begin{eqnarray}
    \label{eq:loss_tg}
    \mathcal{L}_{tg} = \mathcal{L}_{CI} + \mathcal{L}_{F}(H),
\end{eqnarray}
where $H=\{ h_{l} \}_{l=1}^{d}$ are the vectors defined by the transpose of $V$.
The above learning process is summarized as Algorithm\ref{alg:two-stage}.
\begin{algorithm}[t]
    \caption{Two-stage learning in cIDFD.}\label{alg:two-stage}
    \begin{algorithmic}[1]
        \Require
        \Statex dataset $\mathcal{X}=\{ x_{i} \}_{i=1}^{N}$ and $\mathcal{B}=\{ b_{i} \}_{i=1}^{M}$;
        structure of embedding function $f_{\theta}$ and $g_{\psi}$;
        memory banks $\bar{V}=\{ \bar{v}_{i} \}_{i=1}^{N}$, $\bar{W}=\{ \bar{w}_{i} \}_{i=1}^{N}$,
        $\bar{Z}=\{ \bar{z}_{i} \}_{i=1}^{M}$;
        training epochs $E_{f}$ and $E_{g}$;.
        \State Initialize parameters $\theta$, $\psi$, $\bar{V}$, $\bar{W}$, $\bar{Z}$
        \While{$epoch < E_{f}$}\Comment{training for $f_{\theta}$}
        \For{each minibatch $\mathcal{B}_{b}$}
        \State compute $z_{i}=f_{\theta}(b_{i})$ for $b_{i}$ in minibatch $\mathcal{B}_{b}$;
        \State compute the IDFD loss $\mathcal{L}_{bg}$ through Eq.\eqref{eq:loss_bg};
        \State update $\theta$ by the optimizer;
        \State update $\bar{z}_{i}$ by the moving average;
        \EndFor
        \State $epoch \gets epoch + 1$
        \EndWhile
        \While{$epoch < E_{g}$}\Comment{training for $g_{\psi}$}
        \For{each minibatch $\mathcal{X}_{b}$}
        \State compute $v_{i}=g_{\psi}(x_{i})$ and $w_{i}=f_{\theta}(x_{i})$ for $x_{i}$ in minibatch $\mathcal{X}_{b}$;
        \State compute the cIDFD loss $\mathcal{L}_{tg}$ through Eq.\eqref{eq:loss_tg};
        \State update $\psi$ by the optimizer;
        \State update $\bar{v}_{i}$ and $\bar{w}_{i}$ by the moving average;
        \EndFor
        \State $epoch \gets epoch + 1$
        \EndWhile
    \end{algorithmic}
\end{algorithm}

\section{Experiments}
\subsection{Datasets}
\label{sec:dataset}
\begin{figure*}[t]
    \begin{minipage}[b]{0.49\linewidth}
        \centering
        \includegraphics[scale=0.35]{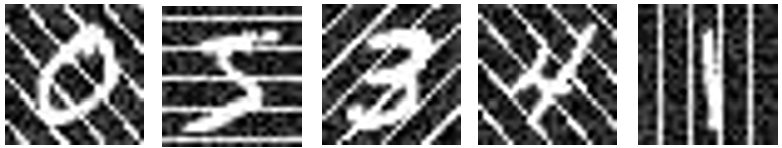}
        \subcaption{Stripe MNIST}\label{fig:sample-smnist}
    \end{minipage}
    \begin{minipage}[b]{0.49\linewidth}
        \centering
        \includegraphics[scale=0.35]{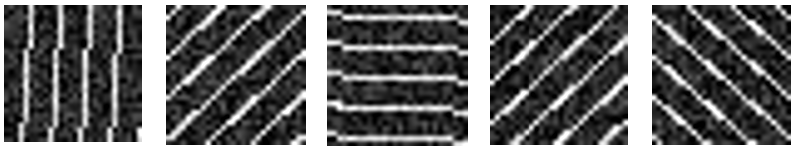}
        \subcaption{Stripe}\label{fig:sample-stripe}
    \end{minipage}
    \begin{minipage}[b]{0.5\linewidth}
        \centering
        \includegraphics[scale=0.35]{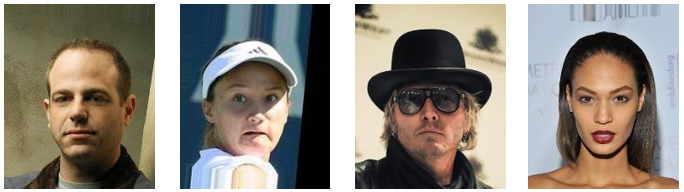}
        \subcaption{CelebA-ROH}\label{fig:sample-celeba-roh}
    \end{minipage}
    \begin{minipage}[b]{0.5\linewidth}
        \centering
        \includegraphics[scale=0.35]{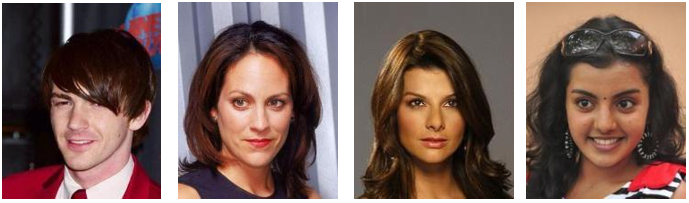}
        \subcaption{CelebA-RNH}\label{fig:sample-celeba-rnh}
    \end{minipage}
    \begin{minipage}[b]{0.5\linewidth}
        \centering
        \includegraphics[scale=0.25]{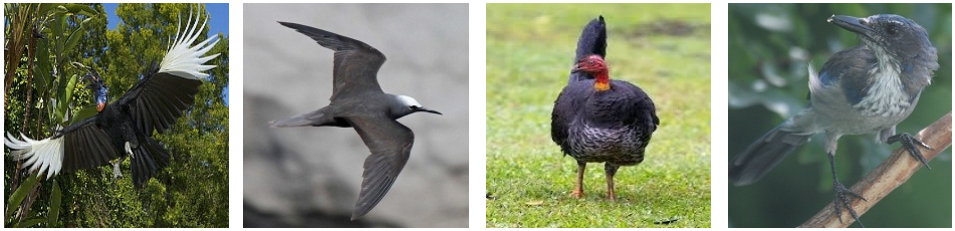}
        \subcaption{Birds400-ABC}\label{fig:sample-birds400}
    \end{minipage}
    \begin{minipage}[b]{0.5\linewidth}
        \centering
        \includegraphics[scale=0.25]{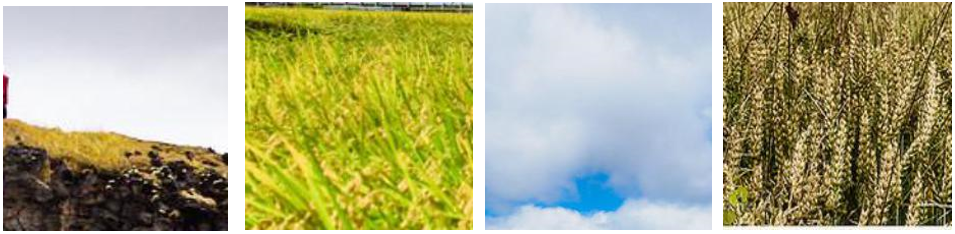}
        \subcaption{Landscape Pictures}\label{fig:sample-landscape}
    \end{minipage}
    \caption{Sample images from each dataset.
        The images on the left are from target datasets, while images on the right are from background datasets.
    }\label{fig:sample-dataset}
\end{figure*}

We evaluated the performance of cIDFD on three datasets created from commonly used public datasets.
Each dataset contains both target and background datasets.
Table~\ref{tab:dataset} summarizes the key details.
Figure~\ref{fig:sample-dataset} shows sample images from each dataset.

\textbf{Stripe MNIST}
We created a synthetic image dataset using handwritten digits from the Modified National Institute of Standards and Technology (MNIST) database~\cite{deng2012mnist} and randomly generated artificial stripe patterns, which are shown in Fig.~\ref{fig:sample-smnist}.
Our goal was to cluster the ten handwritten digits independently of the background patterns.
We also prepared a background dataset of stripe-pattern images that are almost the same as those in the target dataset, but not used in its creation (Fig.~\ref{fig:sample-stripe}).

\textbf{CelebA-ROH}
We assume a clustering task that focuses on certain features of facial images.
We made datasets from the popular celebrity facial images dataset CelebA~\cite{liu2015faceattributes}.
As a target dataset, we collected images with the target attributes "receding hairlines" or "wearing hats" (Fig.~\ref{fig:sample-celeba-roh}).
We used the remaining celebrities as the background dataset, which we call CelebA-RNH (Fig.~\ref{fig:sample-celeba-rnh}).
Our goal was to cluster the two target attributes independently of other attributes
such as eyeglasses, hair color, or gender.

\textbf{Birds400-ABC}
As a target dataset, we collected images from Birds400~\cite{birds}, which are derived from the Kaggle datasets.
Birds400 contains images of 400 types of birds with various backgrounds.
From the training split of the original dataset, we utilized 144 bird species with names starting with A, B, or C (Fig.~\ref{fig:sample-birds400}).
To realize \textit{bird-oriented} clustering by cIDFD, we used the Landscape Pictures dataset~\cite{landsc}, which includes high-quality images of natural landscapes.
By randomly cropping and resizing those images, we generated 41,733 samples with $224 \times 224$ pixels for the background dataset (Fig.~\ref{fig:sample-landscape}).

\begin{table}[t]
    \caption{Image datasets used in the experiments.}
    \label{tab:dataset}
    \centering
    \begin{tabular}{lcccc}
        \toprule
        Dataset            & Image size                  & Samples & classes \\
        \midrule\midrule
        Stripe MNIST       & 28 $\times$ 28 $\times$ 1   & 60000   & 10      \\
        Stripes            & 28 $\times$ 28 $\times$ 1   & 10000   & 4       \\
        \midrule[0.05pt]
        CelebA-ROH         & 178 $\times$ 218 $\times$ 3 & 21065   & 2       \\
        CelebA-RNH         & 178 $\times$ 218 $\times$ 3 & 20000   & -       \\
        \midrule[0.05pt]
        Birds400-ABC       & 224 $\times$ 224 $\times$ 3 & 20822   & 144     \\
        Landscape Pictures & 224 $\times$ 224 $\times$ 3 & 41733   & -       \\
        \bottomrule
    \end{tabular}
\end{table}

\subsection{Comparison with conventional methods}
\label{sec:comparison}
We compared cIDFD with VAE, cVAE and eight other competitive deep clustering methods: CC, MiCE, SCAN, RUC, NNM, TCL, SPICE, IDFD.
Given that VAE and the eight competitive deep clustering methods have no way to handle the background dataset, only the target dataset was used.
In the clustering phase, we applied simple $k$-means to representations for VAE, cVAE, IDFD, and cIDFD.
The number of clusters $k$ is set equal to the number of classes  in each dataset as shown in Table~\ref{tab:dataset}.
Clustering performance was evaluated by three popular metrics: clustering accuracy (ACC), normalized mutual information (NMI), and the adjusted rand index (ARI).
These metrics give values in the range $[0, 1]$, with higher scores indicating more accurate clustering assignments.

Table~\ref{tab:metrics} lists performances for each method.
These results show that cIDFD clearly outperform the conventional methods.
In terms of ACC, the cIDFD scores were improved by approximately 24\% for Stripe MNIST, by 7\% for CelebA-ROH, and by 25\% for Birds400-ABC.
In particular, performance under our method is better than performance under cVAE, which is the most similar method in terms of CA setting usage, even for complex datasets such as CelebA-ROH and Birds400-ABC.

\newcommand{\mc}[2]{\multicolumn{3}{#1}{#2}}
\begin{table*}[t]
    \caption{Clustering results of various methods on three datasets. Results of eight deep clustering methods obtained from our experiments with official code.}
    \label{tab:metrics}
    \centering
    \begin{tabular}{l|ccc|ccc|ccc}
        \toprule
        Dataset        & \mc{c|}{Stripe MNIST} & \mc{c|}{CelebA-ROH} & \mc{c}{Birds400-ABC}                                                                                                       \\
        \midrule
        Metric         & ACC                   & ARI                 & NMI                  & ACC            & ARI            & NMI            & ACC            & ARI            & NMI            \\
        \midrule
        VAE            & 0.177                 & 0.059               & 0.085                & 0.500          & 0.038          & 0.034          & 0.067          & 0.016          & 0.218          \\
        cVAE           & 0.578                 & 0.421               & 0.563                & 0.776          & 0.299          & 0.208          & 0.070          & 0.016          & 0.200          \\
        \midrule
        CC             & 0.377                 & 0.248               & 0.439                & 0.893          & 0.619          & 0.579          & 0.331          & 0.204          & 0.553          \\
        MiCE           & 0.349                 & 0.233               & 0.420                & 0.714          & 0.185          & 0.149          & 0.265          & 0.202          & 0.562          \\
        SCAN           & 0.594                 & 0.505               & 0.696                & 0.577          & 0.023          & 0.021          & 0.257          & 0.152          & 0.479          \\
        RUC            & 0.587                 & 0.504               & 0.706                & 0.482          & 0.019          & 0.016          & 0.005          & 0.031          & 0.310          \\
        NNM            & 0.402                 & 0.321               & 0.548                & 0.632          & 0.070          & 0.055          & 0.226          & 0.140          & 0.470          \\
        TCL            & 0.376                 & 0.233               & 0.421                & 0.820          & 0.409          & 0.456          & 0.332          & 0.193          & 0.547          \\
        SPICE          & 0.492                 & 0.411               & 0.588                & 0.578          & 0.021          & 0.037          & 0.321          & 0.203          & 0.526          \\
        IDFD           & 0.276                 & 0.156               & 0.369                & 0.648          & 0.086          & 0.067          & 0.486          & 0.361          & 0.658          \\
        \midrule
        \textbf{cIDFD} & \textbf{0.830}        & \textbf{0.809}      & \textbf{0.908}       & \textbf{0.969} & \textbf{0.879} & \textbf{0.796} & \textbf{0.738} & \textbf{0.664} & \textbf{0.848} \\
        \bottomrule
    \end{tabular}
\end{table*}

\subsection{Representation distribution}
\label{sec:repdist}
\begin{figure*}[t]
    \begin{minipage}[b]{0.32\linewidth}
        \centering
        \includegraphics[keepaspectratio, scale=0.35]{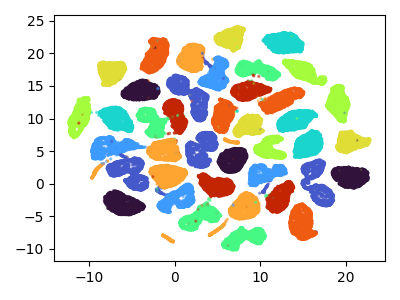}
        \subcaption{Stripe MNIST(IDFD)}\label{fig:idfd_smnist}
    \end{minipage}
    \begin{minipage}[b]{0.32\linewidth}
        \centering
        \includegraphics[keepaspectratio, scale=0.35]{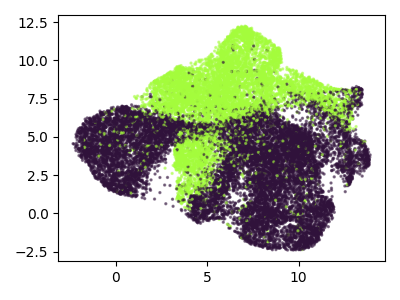}
        \subcaption{CelebA-ROH(IDFD)}\label{fig:idfd_celeba}
    \end{minipage}
    \begin{minipage}[b]{0.32\linewidth}
        \centering
        \includegraphics[keepaspectratio, scale=0.35]{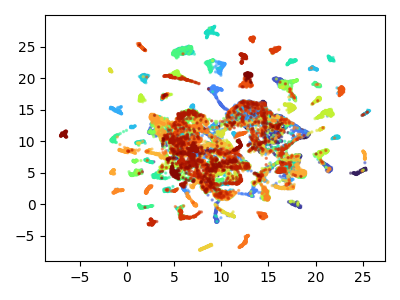}
        \subcaption{Birds400-ABC(IDFD)}\label{fig:idfd_birds}
    \end{minipage}\\
    \begin{minipage}[b]{0.32\linewidth}
        \centering
        \includegraphics[keepaspectratio, scale=0.35]{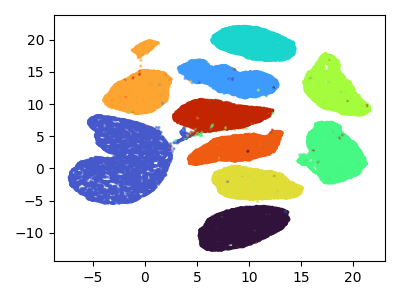}
        \subcaption{Stripe MNIST(cIDFD)}\label{fig:cidfd_smnist}
    \end{minipage}
    \begin{minipage}[b]{0.32\linewidth}
        \centering
        \includegraphics[keepaspectratio, scale=0.35]{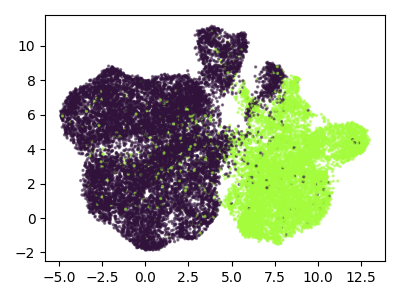}
        \subcaption{CelebA-ROH(cIDFD)}\label{fig:cidfd_celeba}
    \end{minipage}
    \begin{minipage}[b]{0.32\linewidth}
        \centering
        \includegraphics[keepaspectratio, scale=0.35]{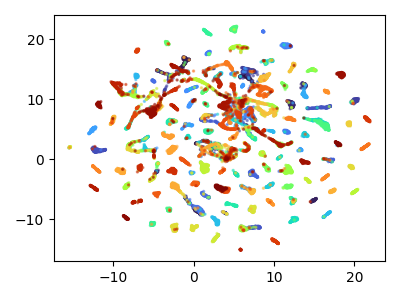}
        \subcaption{Birds400-ABC(cIDFD)}\label{fig:cidfd_birds}
    \end{minipage}
    \caption{Visualizations of feature representations on the Stripe MNIST(10 classes), CelebA-ROH(2 classes), and Birds400-ABC(144 classes) datasets.}\label{fig:reps}
\end{figure*}

Fig.~\ref{fig:reps} visualizes feature representations of the three datasets, which are learned by IDFD and cIDFD.
128-dimensional representations were embedded into two-dimensional space by UMAP~\cite{2018arXivUMAP}.
Point colors indicate ground truth labels.
The distribution clearly shows that cIDFD is preferable to IDFD when grouping samples into clusters characterized by ground truth labels, which are the features we focus on.
For Stripe MNIST, IDFD created clusters excessively according to both the digits and stripes features, while cIDFD generated almost ten clusters.
For CelebA-ROH, IDFD generated one distribution mixing two classes, but cIDFD generated distinct distributions according to the ground truth labels.
cIDFD also correctly distinguished an extremely large number of bird species; however, in the IDFD results, samples of many classes were degenerated to several large clusters.

\subsection{Similarity distribution}
\label{sec:demo}
\begin{figure*}[t]
    \begin{minipage}[b]{0.5\linewidth}
        \centering
        \includegraphics[keepaspectratio, scale=0.3]{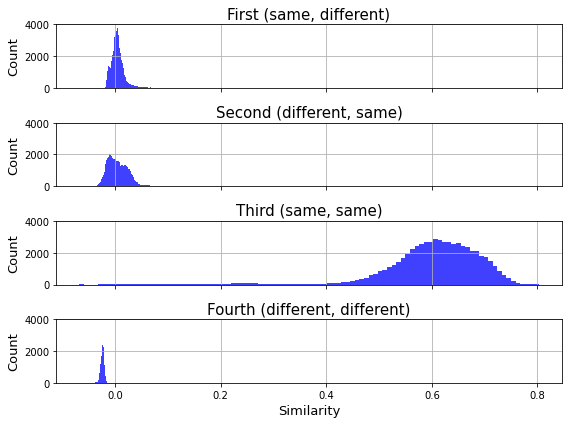}
        \subcaption{IDFD}\label{fig:idfd_simhist}
    \end{minipage}
    \begin{minipage}[b]{0.5\linewidth}
        \centering
        \includegraphics[keepaspectratio, scale=0.3]{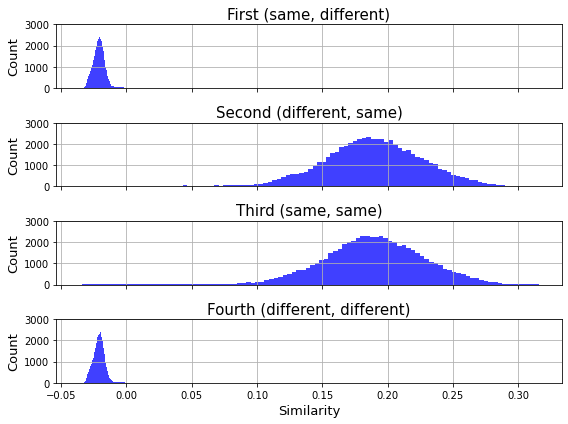}
        \subcaption{cIDFD}\label{fig:cidfd_simhist}
    \end{minipage}
    \caption{Histograms of similarity between samples in four different types of pairs. The pair types are indicated at the top of each figure. For example, the distributions of the first pairs are titled as (same, different), meaning that similarity is calculated for instances of the same background and different target features.
    }\label{fig:simhist}
\end{figure*}

To clearly understand how our method works, we conducted experiments on similarity distribution for both IDFD and cIDFD with the synthetic image dataset Stripe MNIST.
Fig.~\ref{fig:simhist} shows the resulting histograms on four types of average similarity, which were calculated for each instance: the first type is similarity to instances of the same background and a different target; the second type is to instances of a different background and the same target; the third type is to instances of the same background and the same target; and the fourth type is to instances of a different target and a different background.
In the case of IDFD (Fig.~\ref{fig:idfd_simhist}), the peaks of the first and second type of similarity are located in almost same lower region.
As mentioned in Section~\ref{sec:method_ci}, this situation is problematic.
In contrast, cIDFD successfully moved the peak of the second type to the same position of the third type distribution in higher region (Fig.~\ref{fig:cidfd_simhist}).
On the other hand, the first and fourth type distributions are located in the same lower region.
Consequently, instances with the same target features were clustered independently of the background features.

\section{Conclusion}
We presented cIDFD as a new self-supervised clustering method combining instance discrimination with feature decorrelation and contrastive analysis.
Our method is designed to extract unimportant features from a background dataset and reject them in the learning process for the target dataset, resulting in clustering according to only the important features.
The experimental results on the Stripe MNIST, CelebA-ROH, and Birds400-ABC datasets showed that cIDFD outperforms the state-of-the-art SSL method and similar conventional methods with contrastive analysis.
Problem settings allowing utilization of background datasets appear in various fields, so we expect there to be many situations in which cIDFD can be applied.

\bibliographystyle{splncs04}
\bibliography{deep_clustering,self_supervised,contrastive_analysis,ssrl_clustering,others, practical}

\begin{thebibliography}{10}
\providecommand{\url}[1]{\texttt{#1}}
\providecommand{\urlprefix}{URL }
\providecommand{\doi}[1]{https://doi.org/#1}

\bibitem{abid_exploring_2018}
Abid, A., Zhang, M.J., Bagaria, V.K., Zou, J.: Exploring patterns enriched in a
  dataset with contrastive principal component analysis. Nat Commun
  \textbf{9}, ~2134 (2018)

\bibitem{abid_contrastive_2019}
Abid, A., Zou, J.: Contrastive {Variational} {Autoencoder} {Enhances} {Salient}
  {Features}. arXiv:1902.04601  (2019)

\bibitem{caron_unsupervised_2020}
Caron, M., Misra, I., Mairal, J., Goyal, P., Bojanowski, P., Joulin, A.:
  Unsupervised {Learning} of {Visual} {Features} by {Contrasting} {Cluster}
  {Assignments}. In: NeurIPS. pp. 9912--9924. Curran Associates, Inc. (2020)

\bibitem{chen_simple_2020}
Chen, T., Kornblith, S., Norouzi, M., Hinton, G.: A {Simple} {Framework} for
  {Contrastive} {Learning} of {Visual} {Representations}. In: ICML. pp.
  1597--1607. PMLR (2020)

\bibitem{chen_simsiam_2021}
Chen, X., He, K.: Exploring simple siamese representation learning. In: CVPR.
  pp. 15750--15758. IEEE (2021)

\bibitem{Dang_2021_CVPR}
Dang, Z., Deng, C., Yang, X., Wei, K., Huang, H.: Nearest neighbor matching for
  deep clustering. In: CVPR. pp. 13693--13702 (June 2021)

\bibitem{deng2012mnist}
Deng, L.: The mnist database of handwritten digit images for machine learning
  research. IEEE Signal Processing Magazine  \textbf{29}(6),  141--142 (2012)

\bibitem{dirie_contrastive_2019}
Dirie, A.H., Abid, A., Zou, J.: Contrastive multivariate singular spectrum
  analysis. In: 2019 57th Annual Allerton Conference on Communication, Control,
  and Computing (Allerton). pp. 1122--1127 (2019)

\bibitem{ge_rca_2016}
Ge, R., Zou, J.: Rich component analysis. In: ICML. p. 1502–1510. PMLR (2016)

\bibitem{li_contrastive_2021}
Li, Y., Hu, P., Liu, Z., Peng, D., Zhou, J.T., Peng, X.: Contrastive
  {Clustering}. In: AAAI. pp. 8547--8555. AAAI Press (2021)

\bibitem{liu2015faceattributes}
Liu, Z., Luo, P., Wang, X., Tang, X.: Deep learning face attributes in the
  wild. In: Proceedings of International Conference on Computer Vision (ICCV)
  (2015)

\bibitem{2018arXivUMAP}
{McInnes}, L., {Healy}, J., {Melville}, J.: {UMAP: Uniform Manifold
  Approximation and Projection for Dimension Reduction}. arXiv:1802.03426
  (2018)

\bibitem{niu_spice_2022}
Niu, C., Shan, H., Wang, G.: {SPICE}: {Semantic} {Pseudo}-labeling for {Image}
  {Clustering}. arXiv:2103.09382  (2022)

\bibitem{park2021improving}
Park, S., Han, S., Kim, S., Kim, D., Park, S., Hong, S., Cha, M.: Improving
  unsupervised image clustering with robust learning. In: CVPR (2021)

\bibitem{birds}
Piosenka, G.: Birds 400 - species image classification  (2022)

\bibitem{landsc}
ROUGETET, A.: Landscape pictures  (2020)

\bibitem{mizukami}
Shota, M., Yukako, T.: Application of contrastive representation learning to
  unsupervised defect classification in semiconductor manufacturing. In:
  AEC/APC Symposium Asia 2021 (2021)

\bibitem{tao_clustering-friendly_2022}
Tao, Y., Takagi, K., Nakata, K.: {Clustering-friendly} {Representation}
  {Learning} via {Instance} {Discrimination} and {Feature} {Decorrelation}. In:
  ICLR (2021)

\bibitem{mice}
Tsai, T.W., Li, C., Zhu, J.: Mice: Mixture of contrastive experts for
  unsupervised image clustering. In: ICLR (2021)

\bibitem{gansbake_scan_2020}
Van~Gansbeke, W., Vandenhende, S., Georgoulis, S., Proesmans, M., Van~Gool, L.:
  {SCAN}: Learning to classify images without labels. In: ECCV. pp. 268--285.
  Springer International Publishing (2020)

\bibitem{wu_unsupervised_2018}
Wu, Z., Xiong, Y., Yu, S.X., Lin, D.: Unsupervised {Feature} {Learning} via
  {Non}-{Parametric} {Instance} {Discrimination}. In: CVPR. pp. 3733--3742.
  IEEE (2018)

\bibitem{li2022tcl}
Yunfan, L., Mouxing, Y., Dezhong, P., Taihao, L., Jiantao, H., Xi, P.: Twin
  contrastive learning for online clustering. International Journal of Computer
  Vision  (2022)

\bibitem{zhou2022comprehensive}
Zhou, S., Xu, H., Zheng, Z., Chen, J., li, Z., Bu, J., Wu, J., Wang, X., Zhu,
  W., Ester, M.: A comprehensive survey on deep clustering: Taxonomy,
  challenges, and future directions (2022)

\end{thebibliography}

\end{document}